\documentclass{article}

\usepackage{PRIMEarxiv}

\usepackage[utf8]{inputenc} 
\usepackage[T1]{fontenc}    
\usepackage{hyperref}       
\usepackage{url}            
\usepackage{booktabs}       
\usepackage{amsfonts}       
\usepackage{nicefrac}       
\usepackage{microtype}      
\usepackage{lipsum}
\usepackage{fancyhdr}       
\usepackage{graphicx}       
\usepackage{tabularx}
\usepackage{multirow,makecell}
\usepackage{wrapfig}
\usepackage{amssymb}
\usepackage{pifont}
\usepackage{enumitem}
\graphicspath{{media/}}     
\newcommand{\cmark}{\ding{51}}%
\newcommand{\xmark}{\ding{55}}%
\title{LogAI: A Library for Log Analytics and Intelligence
}

\author{
  Qian Cheng, Amrita Saha, Wenzhuo Yang, Chenghao Liu, Doyen Sahoo, Steven Hoi \\
  Salesforce AI Research \\
  \texttt{\{qcheng, amrita.saha, wenzhuo.yang, chenghao.liu, dsahoo, shoi\}@salesforce.com} \\
}

\begin{document}
\maketitle

\begin{abstract}

Software and System logs record runtime information about processes executing within a system. These logs have become the most critical and ubiquitous forms of observability data that help developers understand system behavior, monitor system health and resolve issues. However, the volume of logs generated can be humongous (of the order of petabytes per day) especially for complex distributed systems, such as cloud, search engine, social media, etc. This has propelled a lot of research on developing AI-based log based analytics and intelligence solutions that can process huge volume of raw logs and generate insights. In order to enable users to perform multiple types of AI-based log analysis tasks in a uniform manner, we introduce \textit{LogAI}  (\url{https://github.com/salesforce/logai}), a one-stop open source library for log analytics and intelligence. LogAI supports tasks such as log summarization, log clustering and log anomaly detection. It adopts the OpenTelemetry data model, to enable compatibility with different log management platforms. LogAI provides a unified model interface and provides popular time-series, statistical learning and deep learning models. Alongside this, LogAI also provides an out-of-the-box GUI for users to conduct interactive analysis. With LogAI, we can also easily benchmark popular deep learning algorithms for log anomaly detection without putting in redundant effort to process the logs. We have opensourced LogAI to cater to a wide range of applications benefiting both academic research and industrial prototyping. 
\end{abstract}


\section{Introduction}
\label{sec:intro}
System and Software logs are text messages that are embedded by software and application developers in the source code and are designed to carry useful runtime information about the process, which are typically dumped as raw log files, once the system starts executing.
In modern computer systems, especially for large distributed systems that run complex software, such as search engines, social network websites, and cloud platforms, logs are one of the most critical observability data. Logs are widely used in a variety of operational tasks, covering use cases such as system availability, reliability and security. In scenarios when users have no direct access to the physical servers, logs are often the ground truth about the systems and applications. As such, Log Management has become a very important task in the industrial landscape. In fact, log management market size grew to \$2.29 billion in 2023, at a compound annual growth rate (CAGR) of 15.9\%, according to the report from The Business \cite{Company2023}.

Ideally, logs should be capturing the runtime information at a very granular level and stored permanently so that when any disruptive incident occurs, developers and operators can always look up the correct log file and inspect the log messages to debug what caused the incident. In reality though, because of the colossal size of the log dumps, storing them permanently in the raw form is often impractical. This challenge can be mitigated with the help of large cloud-based logging systems such as AWS Cloudwatch and Microsoft Azure Logs where it is possible to even store the entire log data and retain them for a substantial period of time. 
Moreover, these logging systems also provide capabilities to help efficient log querying and visualization, enabling developers and operators to quickly access the log dumps or log streams of their software. 
With these capabilities, the main open question is, \emph{how to explore raw logs and find the right set of logs associated with an incident?} followed by a more advanced one - \emph{Is there a way to automatically analyze the logs and tell if there are issues with a system, create incidents and provide additional insights?}

Depending on which operational stage logs are involved in, the goal of log analysis in that specific situation could be different. Logs can be used for incident detection, where reliability engineers and developers need to continuously monitor the log streams in order to detect any unexpected behavior that might be indicative of an incident. For post incident detection, log data can play a critical role in root-cause analysis, where operators examine the raw logs to identify the loglines that show anomalous patterns and thus localize the anomaly and eventually the root cause of the incident to a single service, component or module or a group of them. The situation becomes even more complex in large distributed systems, where people (typically reliability engineers) who inspect the logs to resolve incidents may not necessarily be the same group of people (i.e. software and application developers) who write the logging statements in software code. In these situations, understanding even simple dump logs can take significant amount of time and effort, owing to the open-ended nature of the log data. 

Over the past decade there have been various effort targeted at developing both commercial and open-source software to cater to automated log analysis. Though, most of the initial work used either domain specific rules or heuristics, with the proliferation of AI and ML, more and more data-driven techniques have been adopted and popularized in this community. However, most of the AI-driven effort has been applied in an isolated manner, focusing on specific log analysis tasks (like how to extract structure out of the raw logs or how to detect anomaly patterns in it). There is still an urgent need for bringing together all the AI, ML and NLP techniques to a unified platform that can cater to the entire suite of different log analysis tasks. Nevertheless, creating such a one-stop library to serve a diverse set of log-based analytics can be quite non-trivial, with some of the potential challenges being, as follows:   

\begin{itemize}
\item \textbf{Lack of unified log data model for log analysis}. Different logs are in different formats and as a result analysis tools need to be customized for different log formats and schemas. It is not easy to generalize analytical algorithms without a unified data model that can handle heterogenous forms of log data.

\item \textbf{Redundant effort in data preprocessing and information extraction}. The current status of log analytics in this community is that there is a lack of a consolidated pipeline for data preprocessing and information extraction across all log analysis models and tasks - i.e. different log analysis algorithms have been implemented independently, with each adopting their own pipelines and workflows. 
For different tasks, or even different algorithms of the same task, developers need to implement multiple redundant preprocessing and information extraction process modules.

\item \textbf{Difficulty in managing log analysis experiments and benchmarking}. Empirical benchmarking forms a critical part of research and applied science. In the existing literature, there is no unified workflow management mechanism to run log analysis benchmarking experiments. For example, while there has been some isolated pockets of deep learning research for log anomaly detection, it is quite challenging for other organizations or users to adopt them or reproduce their experimental results, due to the lack of a common unified framework for log analysis. 
\end{itemize}

In this inter-disciplinary community of AIOps, users may have different needs while working on log analysis in academic and industrial settings when they are in different roles. For example, 1) Machine learning researchers may need a hassle-free way to perform benchmarking experiments on public log datasets and reproduce the experimental results from peer research groups in order to develop new log analysis algorithms; 2) Industrial data scientists and AIOps practitioners may need an intuitive workflow to quickly experiment with existing log analysis algorithms on their own log data and select the best performing algorithm, hyperparameters and experimental configurations as their log analysis solution, and 3) Data and software engineers need to integrate the selected algorithm into production and deploy them in a smooth and efficient way. Unfortunately, we realize there is no existing open source toolkit that can satisfy all the above needs. 

We are thus motivated to develop a holistic LogAI solution - a python library aimed for conducting AI-based log analytics and intelligence tasks to serve a variety of academic and industrial use-cases. LogAI (\url{https://github.com/salesforce/logai}) provides a unified way to conduct various of log analysis tasks such as log summarization, clustering, anomaly detection. LogAI also provides a unified data model, inheriting from OpenTelemetry log data model, to handle logs in different formats. LogAI is also the first open source log analytics library that incorporate time-series algorithms, statistical learning algorithms and deep learning algorithms. Moreover, LogAI implemented an out-of-the-box GUI portal to conduct log analysis in interactive way, more naturally align with the user experience of real-world log analysis. 

Besides, in this technical report we also demonstrate how to use LogAI to easily benchmark deep learning algorithms for log anomaly detection without any redundant effort in log preprocessing and cleaning. In this community, there are existing libraries like LogLizer and Deep-Loglizer \cite{He2016,DBLP:journals/corr/abs-2107-05908} which have consolidated some of the AI/ML effort for the log domain. However, they still suffer from a few limitations - for example lacking a unified data processing pipeline that is generic across all tasks or algorithms or catering to only anomaly detection as the log analysis task or covering only a specific types of algorithms. In Section \ref{sec:experiments}, we elaborate on the limitations of these existing libraries and also show how LogAI provides a more intuitive framework for designing and managing the experimental settings while performing comparable to Deep-Loglizer.

\section{Related Work}
\label{sec:related}

\begin{figure}[htbp]
    \centering
    \includegraphics[scale=0.8]{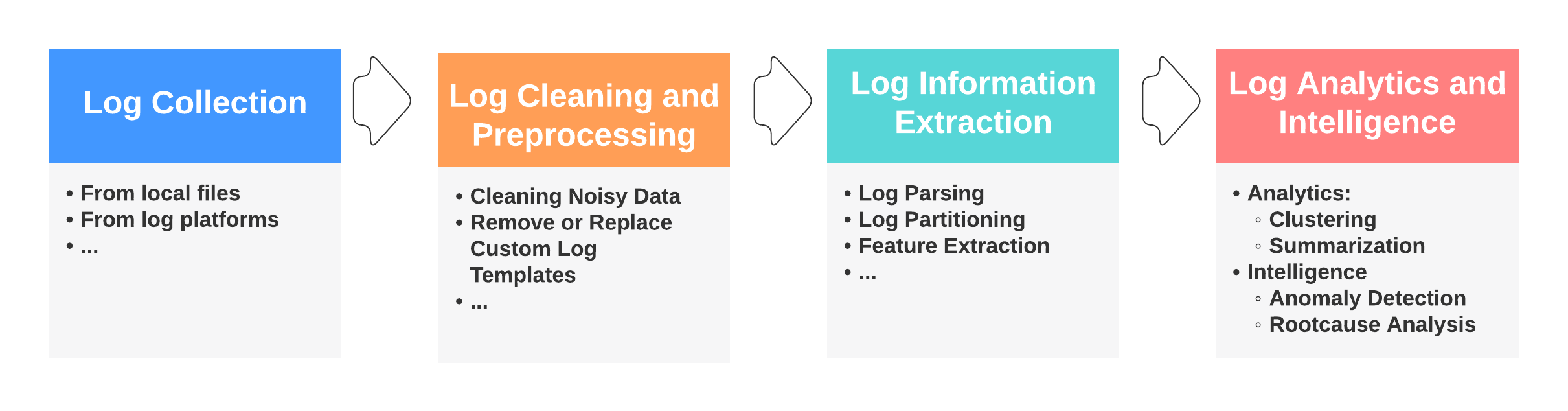}
    \caption{Common Log Analytics and Intelligence Workflow. The common workflow contains four steps: 1) log collection from local log files or log platforms, 2) log cleaning and preprocessing, 3) log information extraction and 4) log analytics tasks (such as clustering and summarization) and log intelligence tasks (such as anomaly detection and root-cause analysis).}
    \label{fig:log_workflow}
\end{figure}

Recently, researchers and engineers have been working on a variety of problems about automated log analysis in academia and industry \cite{Zhaoxue2022}. Based on the existing solutions, we can summarize a common workflow to conduct automated log analysis. The common workflow contains four steps: log collection, log cleaning and preprocessing, log information extraction and log analysis and intelligence applications, Figure \ref{fig:log_workflow}. Log collection is the data loading step that collects logs from local log dump files or log management platforms. Log cleaning and preprocessing is the step to use predefined rules and domain knowledge to clean noisy log data, remove or replace known log templates. This step usually does not involve any ML process. Log information extraction is the step where ML models are involved to extract information from log data, and feed the log representation or features to train ML models for analytics and intelligence application tasks. Log information extraction usually contains several steps like log parsing, partitioning, feature extraction, etc. The final step, log analytics and intelligence, is to train ML models for a specific log downstream task. For example, log clustering and summarization are common log analytics tasks, while log based anomaly detection and root-cause analysis are common log intelligence tasks.

Log analysis has a very long history and there are a lot of tools for log analysis. Almost all commercial log management software/SaaS have associated log analysis/ log insights offerings. This includes log management products such as Splunk, DataDog, NewRelic, etc., as well as cloud providers such as Amazon AWS, Microsoft Azure and Google Cloud. In open source community, there are also very popular log management and analysis projects such as GreyLogs, Grafana, Prometheus, etc. However, neither these commercial log management platform nor open-source log analysis tools are incorporated with comprehensive AI techniques such as deep learning, large language models (LLM), BERT, etc. 

Meanwhile, there are a few open-source AI-based log analysis tools that started to support more comprehensive AI techniques. For example, LOGPAI (https://github.com/logpai/) is one of the most famous log anaysis community on GitHub. LOGPAI provides logparser for automated log parsing. LOGPAI also provides loglizer \cite{DBLP:conf/issre/HeZHL16} and deep-loglizer \cite{https://doi.org/10.48550/arxiv.2107.05908} for log anomaly detection. Besides LOGPAI, there are other open-source projects, most of which are open source code from research outcomes, such as LogClass and Log2Vec from NetManAIOps (https://github.com/orgs/NetManAIOps).

\section{Design Principles}
\label{sec:design}
In this section we discuss about the design principles of LogAI library. LogAI provides a unified framework for log analysis. In order to achieve this, LogAI follows the following design principles: 1) high compatibility with data from different log sources, 2) reusable components to avoid reproducing effort, 3) unified setup process for customized applications and 4) easy-to-use GUI for out-of-box interactive log analysis.

\subsection{Compatible with data from different log sources}
One of the attractive qualities of log data is its open-ended form, where developers can design them to capture useful runtime and performance information to any arbitrary level of granularity as per the needs of the application. 
Different software can generate very different logs. Even in the same software, there are different levels of logs, such as service logs, application logs, systems logs, etc. These logs can be in different formats, either structured, semi-structured or unstructured. LogAI takes these factors into consideration and ensures that the data loader can consume and process these heterogeneous types of logs in a seamless way, by converting these logs into log record with unified log data model.   

\subsection{Reusable components to avoid duplicated effort}
As briefly motivated in Sec \ref{sec:intro}, a particular challenge of building log analytics in both academic and industrial settings,  is the lack of an unified framework that allows reusal of data processing and information extraction components across different log analysis tasks, even on the same data source or dataset. 
For instance, engineers and researchers have to build separate pipelines to perform log anomaly detection, log clustering or summarization even to deal with the same log data source. This burden significantly impacts efficiency in every development stage. from experiments, prototyping all the way to productization. Also running multiple pipelines in production increases the system complexity and brings additional operational cost. Thus, building a library that unifies the interface of common components across multiple downstream tasks is necessary to improve efficiency of all stages of log analysis. 

\subsection{Unified setup process for customized applications}
Even for the same application, the design choice behind the log analysis pipeline might have different variations, based on the various needs or limitations of the use-case. 
For example, log anomaly detection may involve different steps in the end-to-end (E2E) workflow. Some may include log parsing, while others might choose to skip this step either due to the computational overhead or simply because the downstream analysis models do not need a well-defined parsed structure. Also, when converting the raw log text data to machine-readable vectors there can be various choices - either to convert log messages into time-series counter vectors or into event sequences by representing each log line as a id or as a sequence of natural language tokens.  In production setup, adding, removing or replacing a component in the E2E workflow could be very time consuming. LogAI is designed to support building customized applications with easy plug-in / plug-out components, enabling users to quickly try out various combinations through simple intuitive mechanisms like configurable json or yaml files. 

\subsection{Easy-to-use GUI for out-of-box interactive log analysis}
Another learning while we work with different types of log data is about visual examination. Unlike many machine learning domains where the model performance evaluation can heavily rely on metrics, such as Precision, Recall, F-scores, log analysis tasks usually need more visual examination to validate the performance. Thus, LogAI is developed with a graphic user interface (GUI), or a portal, to integrate with interactive analytical features for tasks such as log summarization, clustering and anomaly detection. We believe this portal can reduce the cognitive overhead on the LogAI users in onboarding to the library and help them execute the log analysis tasks quickly and intuitively. 

\section{Architecture}
\label{sec:architecture}
LogAI is separated into the GUI module and core library module. The GUI module contains the implementation of a GUI portal that talks to backend analysis applications. The portal is supported using Plotly Dash (https://plotly.com/dash/). The core library module contains four main layers: data layer, pre-processing layer, information extraction layer and analysis layer. Each layer contains the components to process logs in a standard way. LogAI applications, such as log summarization, log clustering, unsupervised log anomaly detection, are created on top of the components of the four layers.

\subsection{Core Library Modules}

\begin{figure}[ht]
    \centering
    \includegraphics[width=0.9\textwidth]{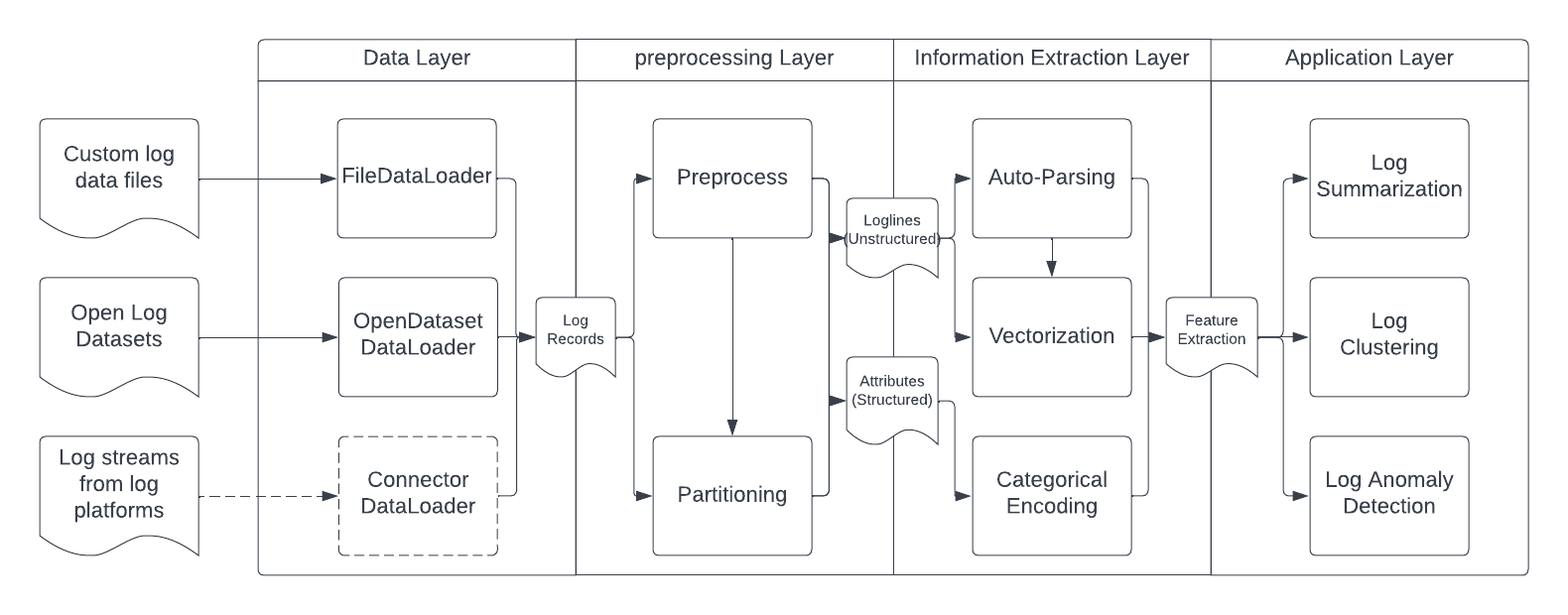}
    \caption{LogAI Architecture}
    \label{fig:architecture}
\end{figure}

LogAI is implemented in the architecture described in Figure \ref{fig:architecture}. In this section we describe the technical details of each layer. Including the implementation of components and how the components communicate across layers.

\subsubsection{Data Layer}
Data layer contains two component classes: \code{LogRecordObject} class and \code{DataLoader} class. 

\code{LogRecordObject} class defines the data model of log records. As we mentioned in Introduction, logs are free-form text and can be unstructured or semi-structured. Even for structured logs, different software applications may name their log data in different ways. \code{LogRecordObject} is to adapt log data from different sources to a more unified structure in order to provide a data object that can be used in all follow-up processes without modification. In LogAI, data model of \code{LogRecordObject} is a subset of the log and event record definition by OpenTelemetry (https://opentelemetry.io/), containing fields in Table \ref{tab:log_record_object}.

\begin{table}[ht]
 \caption{\code{LogRecordObject} Data Model}
  \centering
  \begin{tabularx}{\textwidth}{|l|X|}
    \hline
    Field  & Description \\\hline\hline
    Timestamp & Timestamp when event occurred. \\\hline
    Body & loglines or the content of log messages. \\\hline
    Attributes &  a map<key,value> for structured information of log record. \\\hline
    TraceId  & Request trace id as defined in W3C Trace Context. Can be set for logs that are part of request processing and have an assigned trace id. This field is optional. \\\hline
    SpanId    & Trace flag as defined in W3C Trace Context specification. At the time of writing the specification defines one flag - the SAMPLED flag. This field is optional.  \\\hline
    SeverityText & String represents the severity. This field is optional. \\\hline
    SeverityNumber & Numeric values of severity, TRACE(1-4), DEBUG(5-8), INFO(9-12), WARN(13-16), ERROR(17-20), FATAL(21-24). This field is optional. \\\hline
    Resource  &  Description of. the source of the log. \\\hline
    InstrumentationScope  &  Multiple occurrences of events coming from the same scope can happen across time and they all have the same value of InstrumentationScope. \\\hline
  \end{tabularx}
  \label{tab:log_record_object}
\end{table}

\code{DataLoader} is a class that implements functions to load data from sources. In current version we implement \code{FileDataLoader} to load data from local files, \textit{e.g.} \code{.log},\code{.csv},\code{.tsv},\code{.json}. The associated \code{DataLoaderConfig} class defines the configuration of how data will be loaded. \code{load\_data()} method will load data from target source and return \code{LogRecordObject}. In the future versions we will support data loaders with connectors to consume data directly from log platforms such as Splunk, Datadog, AWS Cloudwatch, etc.

\subsubsection{Preprocessing Layer}

\begin{figure}
    \centering
    \includegraphics[scale=0.7]{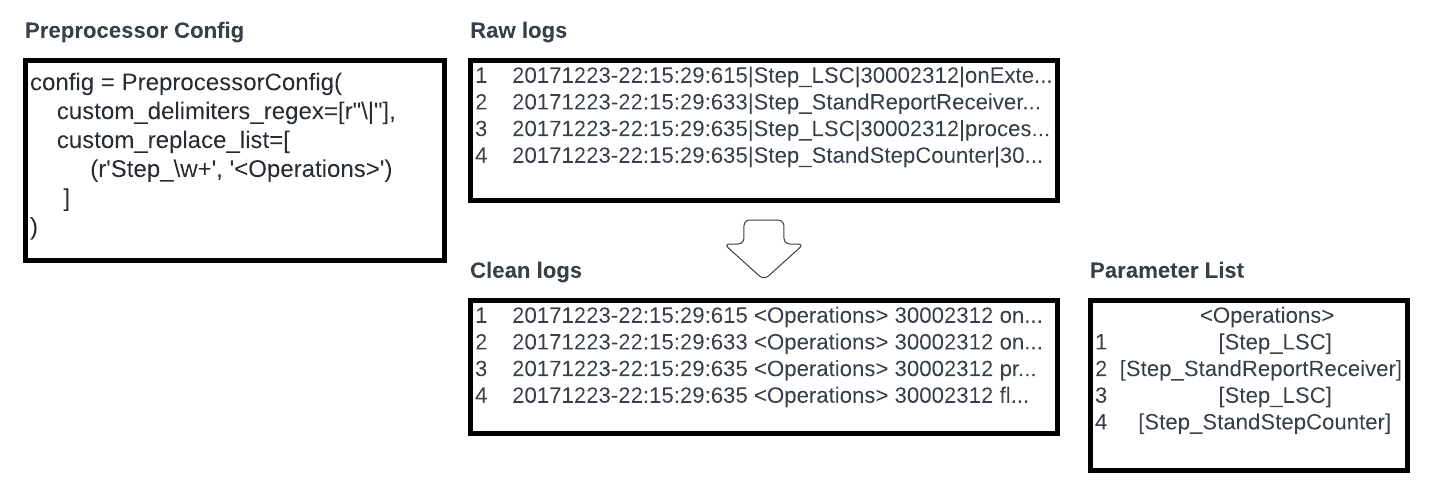}
    \caption{Example of preprocessor execution}
    \label{fig:preprocessor}
\end{figure}

\textbf{Preprocessing.} \code{Preprocessor} is a class to conduct logline level preprocessing. Users can initialize a \code{preprocessor} instance with configuration and execute \code{.clean\_log()} method to obtain cleaned loglines. The supported configuration includes \code{custom\_delimiters\_regex} to parse logs with custom delimiters and \code{custom\_replace\_list} to identify and replace the substrings that match regex patterns in this list, examples are show in Figure \ref{fig:preprocessor}.

\textbf{Partitioning.} \code{Partitioner} is  a class that helps partitioning the logs. As part of the preprocessing, there are needs to shuffle, concatenate and sequentialize raw logs into different forms, for example using time-based partitions or identifier-based partitions or sliding window partitions of fixed lengths. This class provides optional functions for this type of process.

\subsubsection{Information Extraction Layer}
Information extraction layer  contains modules to convert log records into vectors that can be used as input of machine learning models for the actual analytical tasks. Current log analysis research and applications indicate three main input data types are used in the ML approaches: 1) converting log records into counter vectors to use time-series ML techniques, 2) converting log records into feature vectors to use tabular-based ML techniques and 3) converting log records into sequences to use sequential ML techniques. 

LogAI implemented four components in the information extraction layer to extract information from the log records and convert logs to the target formats. Log parser component implements a series of automatic parsing algorithms in order to extract templates from the input loglines. Log vectorizer implements a bag of vectorization algorithms to convert free-form log text into numerical representations for each logline. Categorical encoder implements algorithms that encoding categorical attributes into numerical representations for each logline. Last but not least, feature extractor implements methods to group the logline level representation vectors into log event level representations. 

\textbf{Automated Log Parsing.} \code{LogParser} is a class that conducts automated log parsing tasks. Currently LogAI covers three automated log parsing algorithms:  DRAIN\cite{He2017}, IPLoM\cite{Makanju2009} and AEL\cite{Jiang2008}. LogParser takes the unstructured logline text as input and generate two sets of results: \code{parsed\_logline} are the static pattern of all logs in this category, \code{parameter\_list}  are the lists of values for each “*” position in the log pattern for the same set of loglines.

\textbf{Log Vectorization.} \code{LogVectorizer} is a class that converts unstructured loglines into semantic vectors. Each semantic vector is an array of numeric values that represents this logline text. LogVectorizer supports popular text vectorization algorithms such as \code{TF-IDF} \cite{Ramos2003}, \code{FastText} \cite{Bojanowski2017}, \code{Word2Vec} \cite{Mikolov2013}, etc.

\textbf{Categorical Encoding.} \code{CategoricalEncoder} is a class that encodes log attributes, the structured portion of logs. The string type attributes will be transformed into categorical representations. \code{CategoricalEncoder} supports popular categorical encoding algorithms such as label encoding, one-hot encoding, ordinal encoding etc.

\textbf{Feature Extraction.} \code{FeatureExtractor} is a class that conducts final transformation of raw log data into log feature set that machine learning models can consume. In LogAI, we primarily cover three types of log features: 1) time-series counters, 2) semantic feature sets and 3) sequence vectors. Time-series counters will be used to feed time-series models such as ETS, ARIMA. Semantic feature set can be widely used in a variety of machine learning and deep learning models. Sequence vectors are a specific type of feature format that are required by sequence-modeling based deep learning methods, for example Recurrent Neural Network or Convolutional Neural Networks.

\subsubsection{Analysis Layer}

The analysis layer contains modules that conduct the analysis tasks, including but not limit to semantic anomaly detector, time-series anomaly detector, sequence anomaly detector, clustering, etc. Each analysis module provides unified interface for multiple underlying algorithms. 

\textbf{Anomaly Detection.} \code{AnomalyDetector} is a class to conduct anomaly detection analysis to find abnormal logs from semantic perspective. \code{AnomalyDetector} takes log features of the given logs as input. The output are the anomaly scores. LogAI supports two different types of anomaly detection: 1) anomaly detection based on log counter vectors, 2) anomaly detection based on log semantic representations. The supported anomaly detection algorithms includes univariate and multivariate time-series analysis algorithms from Merlion \cite{Bhatnagar2021}, unsupervised outlier detection models like one-class SVM \cite{10.1162/089976601750264965} and local outlier filter (LOF) \cite{10.1145/342009.335388} from scikit-learn \cite{Pedregosa2011}. 

\textbf{Deep-learning based anomaly detection}. \code{NNAnomalyDetector} class supports deep-learning model based log anomaly detection algorithms, most of which are taking log sequence vectors as input. LogAI integrate some of the popular deep learning based algorithms like recurrent neural network (RNN) based model LSTM \cite{HochSchm97}, convolutional neural network (CNN), Transformers \cite{10.5555/3295222.3295349} and pretrained Transformer based Language Model BERT \cite{DBLP:conf/naacl/DevlinCLT19}. The output are anomaly scores for each log sequence.

\textbf{Clustering.} \code{Clustering} is a class to conduct log clustering analysis tasks. The input for log clustering are the semantic log features. \code{Clustering} is integrated different clustering models, such as k-Means \cite{10.1145/1772690.1772862}, DBSCAN \cite{10.1145/3068335} etc. The output is a map between each log feature record and a cluster label. 

\subsubsection{E2E Applications} 

\begin{figure}[ht]
    \centering
    \includegraphics[width=0.7\textwidth]{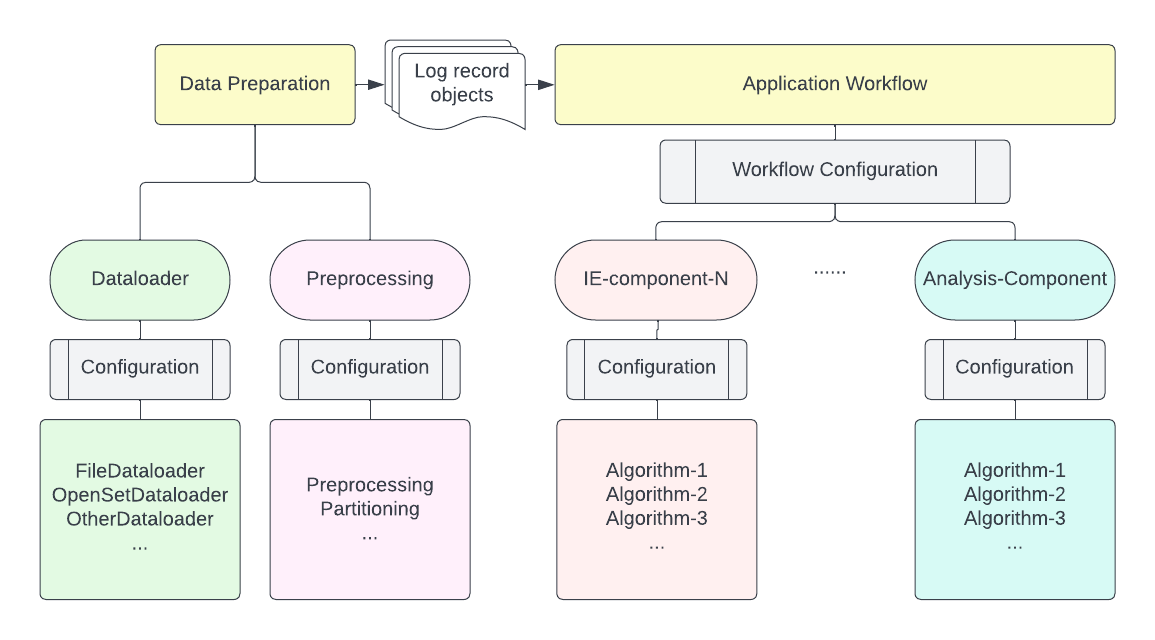}
    \caption{Design Principles of E2E Applications}
    \label{fig:design_principles}
\end{figure}

Depending on the component modules from data layer, preprocessing layer, feature extraction layer and analysis layer, LogAI provides the flexibility to build end-to-end log analysis applications. And the applications follows below design principles \ref{fig:design_principles}. LogAI is launched with several out-of-the-box applications. 

\textbf{Log Summarization.} It is very important to understand your logs before using them for downstream tasks. Log summarization leverages machine learning to process, aggregate and summarize logs. Please refer to the GUI module Section \ref{sec:gui_module} for more detail about how to use.

\textbf{Log Clustering.} Log clustering can be used to categorize logs. Finding meaningful clusters can bring benefits in a variety of use cases like anomaly detection, log storage, query, etc. Please refer to the GUI module Section \ref{sec:gui_module} for more detail about how to use.

\textbf{Log Anomaly Detection.} Log anomaly detection is an application that detect anomalous loglines. Here in LogAI log anomaly detection can detect both time-series anomalies and semantic anomalies. Please refer to the GUI module Section \ref{sec:gui_module} for more detail about how to use.

\subsection{GUI Module}\label{sec:gui_module}

\begin{figure}
    \centering
    \includegraphics[width=0.9\textwidth]{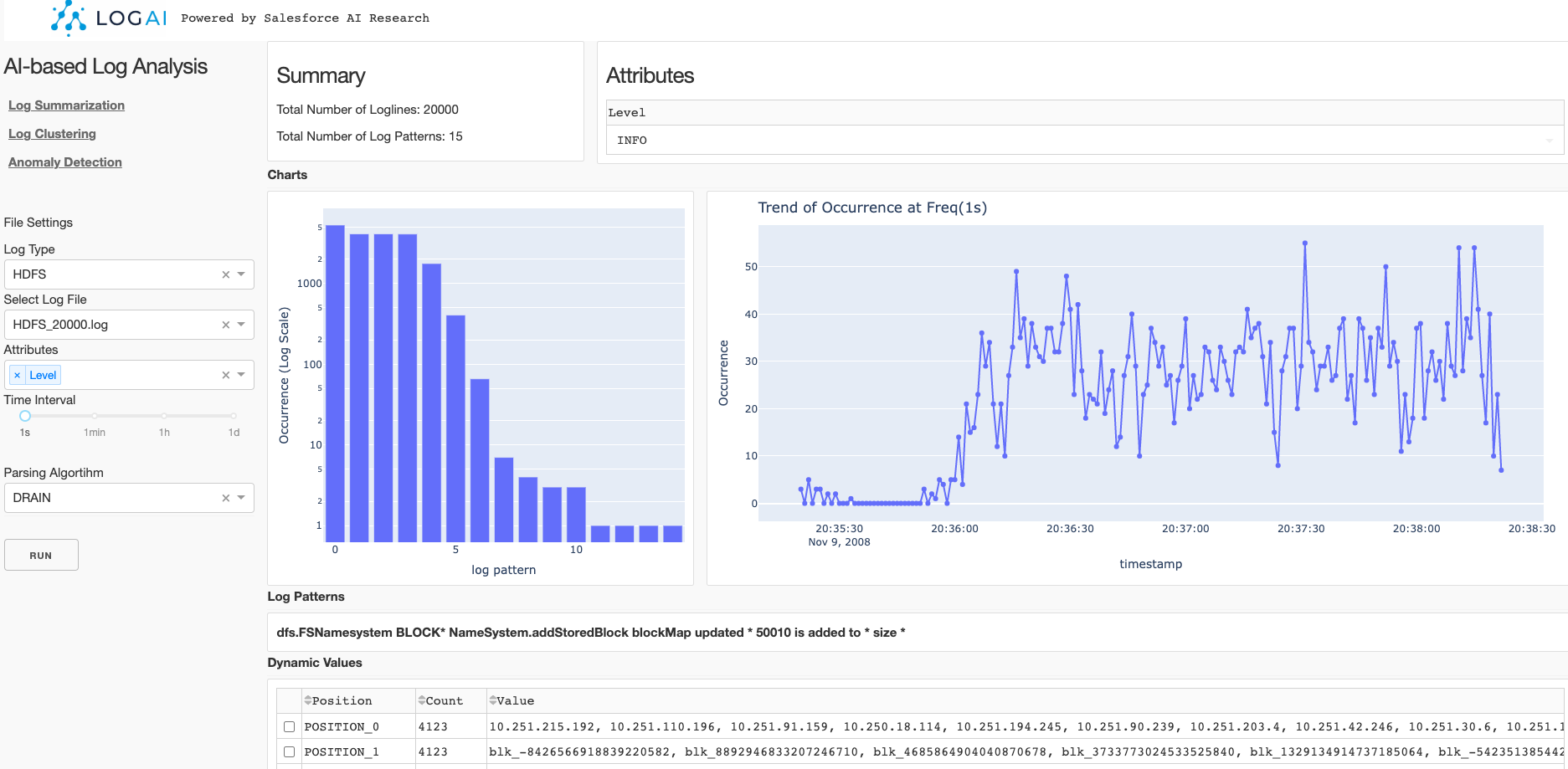}
    \caption{LogAI GUI portal}
    \label{fig:gui}
\end{figure}

The GUI module is implemented to provide a web portal for the out-of-the-box log analysis applications, including log summarization, log clustering and log anomaly detection. Figure \ref{fig:gui} shows the log summarization of LogAI portal. LogAI portal is developed using Plotly Dash framework. 

\textbf{Control Panel.} Control panel is on the left side of the page. In the control panel, users can upload files, configure file and algorithm settings. When the user click "Run" button, the analysis execution is triggered. This behavior is uniform for all three different applications. After analysis execution completed, the results will be displayed on the right side of the page.

\textbf{Main Display Panel.} On the right side of the page we display the analysis results. Different applications may have different layouts. The portal supports interactive visualization. The users can click or hover on parts in the charts to drill down and get more detailed information. 

The interaction between frontend and backend of different applications are designed to be unified. The control panel collects user input and generate configuration for application and send to backend. Backend consumes the configuration to create component instances to execute the workflow. After finishing the job, it will send the result table to frontend. The display panel for each application controls how the result table will be rendered for visualization. Users can expand the GUI portal to support customized analysis applications by following the same design pattern and reusing the existing components.

\subsection{Summary of Supported ML Algorithms in LogAI}

This section summarizes the machine learning algorithms supported in LogAI. LogAI provides an \code{algorithms} component to implement all supported algorithms with algorithm factory. The \code{algorithm} contains five algorithmic modules, notably: parsing\_algo, vectorization\_algo, categorical\_encoding\_algo, clustering\_algo, anomaly\_detection\_algo. \code{algorithms} component also contains a \code{nn\_model} module to implement all neural network models. LogAI has defined unified algorithm interfaces for each module and we can implement more algorithms and integrated it with LogAI in future development. The current LogAI algorithm coverage is shown in Table \ref{tab:sup_algo}.

 The deep-learning models generally being much more parameter-heavy, require more high-end compute devices like GPU. In such cases, their LogAI implementations provide options to use different devices (CPU or GPU) or multiple GPUs seamlessly through the algorithm parameter configurations.

\begin{table}[ht]
\centering
\caption{Summary of supported machine learning algorithms in LogAI}
{\renewcommand{\arraystretch}{1.2}
\begin{tabular}{| p{0.2\textwidth} | p{0.2\textwidth} | p{0.5\textwidth} |}
\hline
Module              & Algorithm & Task \\
\hline\hline
\multirow{3}{*}{Log parser} & DRAIN & \multirow{3}{*}{Information Extraction}    \\
                  &  IPLoM         &       \\
                  &  AEL         &       \\\hline
\multirow{6}{*}{Log vectorizer}  & Fast-text          & \multirow{6}{*}{Unstructured Log Representation}\\
                  &  TF-IDF         &   \\
                  &  Word2Vec         &     \\
                  &  Semantic         &     \\
                  &  Sequential         &     \\  
                  &  BertTokenizer         &   \\\hline
\multirow{3}{*}{Categorical Encoder}  & Label encoding         &  \multirow{3}{*}{Structured Log Representation
}\\
                  &  OneHot Encoding        &   \\
                  &  Ordinal Encoding         &     \\\hline
\multirow{3}{*}{Clustering}  & DBSCAN         &  \multirow{3}{*}{Analysis: Log Clustering}       \\
                  &  K-means       &   \\
                  &  BIRCH         &      \\\hline  
\multirow{7}{*}{Anomaly Detection}  & One-class SVM         &  \multirow{4}{*}{Analysis: Outlier Detection}       \\
                  &  Isolation Forest       &  \\
                  &  LOF       &   \\
                  &  Distribution divergence       &   \\ \cline{2-3}
                  &  ETS       &  \multirow{3}{*}{Analysis: Time-series Anomaly Detection} \\
                  &  Dynamic Baseline       &  \\
                  &  ARIMA         &     \\ \hline 
\multirow{3}{*}{NN models}                  & CNN    &  \multirow{3}{*}{Analysis: Sequential Anomaly Detection} \\
                  &  LSTM      &   \\
                  &  Transformers         &   \\ \cline{2-3}
                  &  LogBERT         &   Analysis: (Sequential / Non-Sequential) Anomaly Detection\\
\hline
\end{tabular}}
\end{table}
\label{tab:sup_algo}

\section{Experiments: Benchmarking Log Anomaly Detection}
\label{sec:experiments}
In this section, we elaborate some of the experimental effort at building pipelines for specific log analysis tasks on publicly available log datasets. The purpose of this is to benchmark the performance of our LogAI library on these standard tasks with the performances reported in existing literature or other well-known log libraries. 

Amongst the different log analysis tasks, log based anomaly detection is perhaps the most objective task, where domain experts like reliability and performance engineers can provide some supervision around which log sequences show anomalous behavior. The other tasks like log clustering, summarization are much more subjective in nature while log based root cause analysis is too specific and tightly coupled with the application or environment it is deployed in. Hence for these tasks it is often impossible to collect supervision labels for benchmarking purposes. Consequently most of the publicly available log analysis datasets and benchmarks have focused on the anomaly detection task. While a small subset of these datasets have also been redesigned to serve log clustering and log summarization in past literature, they can at best be considered as pseudo-oracle data for these tasks and are still are not large-scale enough for benchmarking purposes. Hence, for this reason, in our LogAI library we focus on benchmarking only the log based anomaly detection task.

Following the advances of Artificial Intelligence (AI), Machine Learning (ML) and Natural Language Processing (NLP), for log anomaly detection task also traditional statistical ML based solutions (like SVM, Isolation Forest etc.) have gradually given way to more powerful and sophisticated neural models. Some of these newer models can leverage self-supervised learning to achieve comparable anomaly detection performance in unsupervised settings in comparison to older traditional supervised models. Additionally, the traditional ML models having being around for quite a while, have been more extensively studied with fairly well-reproduced benchmarks in existing literature. Hence in our benchmarking experiments, we have only focused on the more recent neural models.  

\subsection{Limitations of Existing Libraries and Benchmarking Practices} 
Over the past decade have been numerous literature \cite{10.1007/s11036-021-01832-3,10.1145/3460345,10.1145/3483424,10.1145/3459637.3482209,10.1145/3510003.3510155,10.1145/3468264.3473933,DBLP:journals/corr/abs-2112-03159,10.1145/3501297,DBLP:journals/corr/abs-2107-05908} reporting the log anomaly detection performance on some of the standard open-sourced log datasets, as well as various effort at open-sourcing libraries catering the log anomaly detection task. For example, \cite{He2016,DBLP:journals/corr/abs-2107-05908} had released libraries (Loglizer and Deep-Loglizer) for log based anomaly detection using traditional machine learning and more recent deep learning models, respectively. In their library they had consolidated some of the benchmarking effort, bringing together all the popular log anomaly detection models for a more fair comparison on a few public log datasets. 

However, despite this, there is still a lack of rigorous standardisation and benchmarking amongst these works, especially the ones employing neural models. Below we list some of the specific limitations of Loglizer and Deep-Loglizer library which necessitates the need for an unified, generic framework for log analysis tasks: 
\begin{itemize}[leftmargin=*]
    \item \textbf{Generic Log Data Processing Pipeline:} There is a lack of libraries that provide a generic data processing pipeline that is common across different log datasets or different log anomaly algorithms. While Loglizer \cite{DBLP:conf/issre/HeZHL16} and Deep-Loglizer \cite{DBLP:journals/corr/abs-2107-05908} has achieved this to some degree, they still require some dataset-specific preprocessing and customization which are quite open-ended. For users wishing to replicate on their own datasets or other public datasets, there is no clear framework guiding the necessary steps and output-structure of the dataset-specific preprocessing to follow. On the other hand, LogAI library provides a an unified generic data-processing pipeline across all public datasets and log analysis algorithms. It only requires a very minimal dataset-specific customization with a clear template of the kind of preprocessing needed for each dataset - for e.g. each dataset has its own way of specifying the fields of the LogRecordObject (governed by OpenTelemetry data models) e.g. labels or identifiers of the loglines - which are either directly part of the raw log data or have to be derived based on some rules. 
    \item \textbf{Catering to multiple Log Analysis Tasks}: There is a lack of libraries that can cater to all kinds of log analysis tasks (including log clustering, summarization, anomaly detection etc) under a single generic platform. Each of the existing log libraries are tailored for a specific kind of log analysis task. For example libraries like loglizer and Deep-Loglizer specifically focus on log based anomaly detection, log-parser on parsing log data and log3C cater to clustering and correlation specific analysis. On the other hand, logAI enables all of these analysis tasks along with others, like, summarization, visualization etc under an unified framework. 
    \item \textbf{Coverage of Log Analysis Models}: The existing Loglizer library provides the more traditional machine learning algorithms for log based anomaly detection, with the Deep-Loglizer being a deep-learning based counterpart of it, providing only neural ML models. LogAI on the other hand, provides a generic framework encompassing most of the popular AI/ML algorithms - starting from traditional statistical ML models to popular neural models as well as more recent pretrained Transformer (BERT) based models. Going ahead, our logAI library can provide a more extended platform for integrating with more upcoming and powerful neural models as the mainstream deep learning research progresses. For all of these models, logAI provides a single unified data processing platform, that is independent of the kind of downstream analysis task or models. 
\end{itemize}
Thus, with LogAI library, we aim at a more intuitive and easy-to-use log analysis framework for practitioners of different areas and levels of expertise to perform log analysis, without being impeded by the technical nuances of the task.

\subsection{Log based Anomaly Detection Workflow}
\label{subsec:log_ad_workflow}
In order to handle the complex and heterogenous nature of log data, log based anomaly detection typically follows a multi-step pipeline. Starting with the raw log data dump or data streams, the log analysis workflow does some initial preprocessing and cleaning-up of the raw logs to make them amenable to ML models. This is typically followed by log parsing which extracts a loose structure from the semi-structured data and then performs grouping and partitioning of the log lines into log sequences in order to model the sequence characteristics of the data. After this, the logs or log sequences are vectorized i.e. represented as a machine-readable vector, by first tokenizing each instance and converting each token to a $d$-dimensional embedding. On this vectorized version of the log data, various anomaly detection models can be applied.

The choices of each of these steps (for e.g. whether to apply parsing or not, or whether to partition based on sessions or sliding windows, or whether to apply clustering or not) can be guided by various factors - nature of the application generating the log data or the model requirements or other efficiency or performance related constraints. 

\textbf{i) Log Preprocessing:} In LogAI, this step involves handling the formatting of timestamps, logline-identifiers and any associated labels (e.g. anomaly labels) in the raw log data to make it compatible to openTelemetry data. Additionally it also provides customised filtering of specific regular expression patterns (like IP addresses or memory locations or file paths) that are deemed irrelevant for the actual log analysis. 

\begin{wrapfigure}[11]{l}{0.5\textwidth}
    \centering
    \fbox{\includegraphics[width=0.9\linewidth]{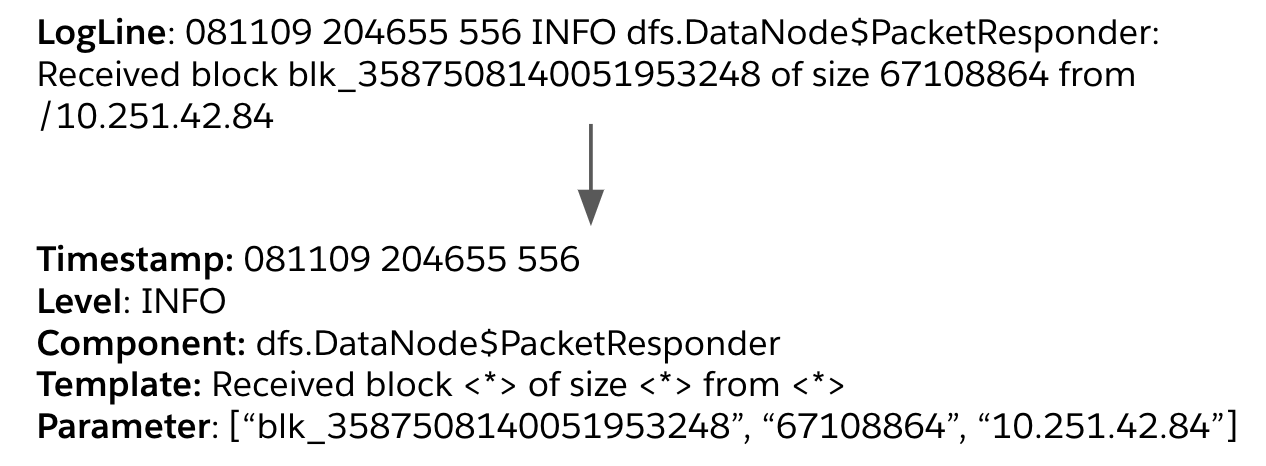}}
    \caption{Example of Log Parsing}
    \label{fig:log_parsing}
\end{wrapfigure}
\textbf{ii)Log Parsing:} To enable downstream processing, unstructured log messages first need to be parsed into a structured event template (i.e. constant part that was actually designed by the developers) and parameters (i.e. variable part which contain the dynamic runtime information). Figure \ref{fig:log_parsing} provides one such example of parsing a logline. In LogAI library we provide three popular log parsers which use heuristic-based techniques - Drain \cite{8029742}, IPLoM \cite{10.1145/1557019.1557154} and AEL \cite{4601543}. 

\vspace{1em}
\begin{wrapfigure}[13]{l}{0.35\textwidth}
    \centering
    \fbox{\includegraphics[width=0.9\linewidth]{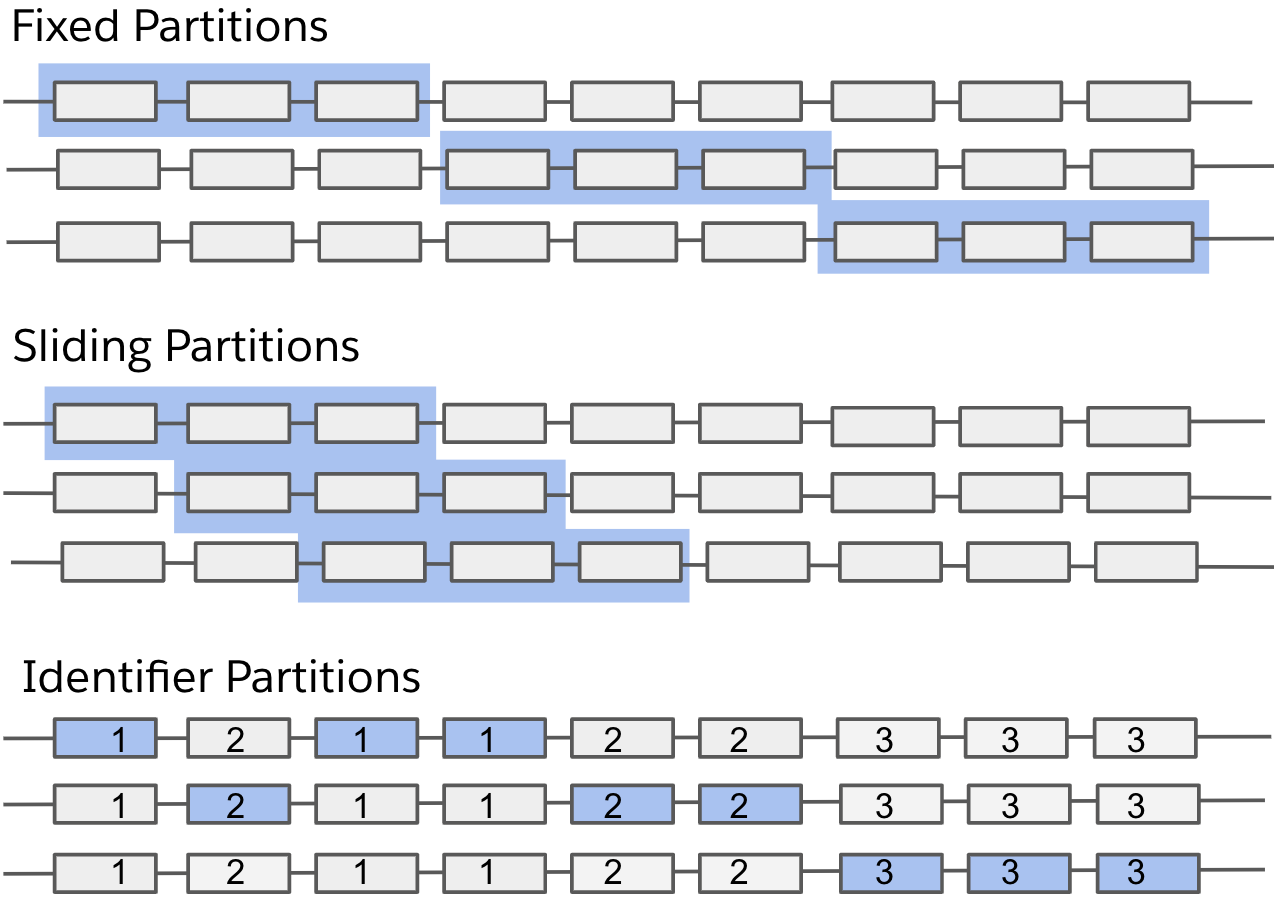}}
    \caption{Different types of log partitioning}
    \label{fig:log_partition}
\end{wrapfigure}

\textbf{iii) Log Partitioning:} After parsing the next step is to partition the log data into groups, based on some semantics where each group represents a finite chunk of log lines or log sequences. The main purpose behind this is to decompose the original log dump, which typically consists of millions of log lines into logical chunks, so as to enable explicit modeling on these chunks and allow the models to capture anomaly patterns over sequences of log templates or log parameter values or both. In literature, various Log partitioning techniques have been applied \cite{10.1145/3468264.3473933,7381796}. In LogAI we provide  different schemes like - Fixed or Sliding window based partitions, where the length of window is determined by length of log sequence or a period of time, and Identifier based partitions where logs are partitioned based on some identifier (e.g. the session or process they originate from). Figure \ref{fig:log_partition} illustrates these different choices of log grouping and partitioning. A log event is eventually deemed to be anomalous or not, either at the level of a log line or a log partition.

\vspace{1em}
\textbf{iv) Log Vectorization:} After log partitioning, the next step is to represent each partition in a machine-readable way (e.g. a vector or a matrix) by extracting features from them. This can be done in various ways \cite{10.5555/3367471.3367702,7381796}. In LogAI we provide the following vectorization techniques - 
\begin{itemize}[leftmargin=*]
    \item i) sequential representation which converts each partition to an ordered sequence of log event ids 
    \vspace{-0.2em}
    \item ii) quantitative representation which uses count vectors, weighted by the term and inverse document frequency information of the log events 
    \vspace{-0.2em}
    \item iii) semantic representation captures the linguistic meaning from the sequence of language tokens in the log events and learns a high-dimensional embedding vector for each token in the dataset. 
    \vspace{-0.2em}
\end{itemize}
The nature of log representation chosen has direct consequence in terms of which patterns of anomalies they can support - for example, for capturing keyword based anomalies, semantic representation might be key, while for anomalies related to template count and variable distribution, quantitative representations are possibly more appropriate. The semantic embedding vectors themselves can be either obtained using pretrained neural language models like GloVe, FastText, pretrained Transformer like BERT, RoBERTa etc. Or they can also be learnt from scratch on the available training data, by building custom vocabulary and using these neural language models.

\vspace{1em}
\textbf{v) Log Anomaly Detection Models for benchmarking:} The task of log based anomaly detection is to analyze a dump of log data, consisting of a series of timestamped log lines and identify the anomalous log lines that are potentially incident-indicating. Based on the kind of application, log anomaly signals can either be used to detect or localize an already occurred incident or disruption or used to forecast future potential faults or failures. In literature, log based anomaly detection models have been broadly categorized into two types - Supervised and Unsupervised, based on the kind of training framework they follow. Since our objective is to benchmark only neural models, we limit our discussion in this section to this class of models alone. 

\textbf{Supervised Anomaly Detection models} require the anomaly label to be available at the level of each log line or a log group or partition. Furthermore, they typically assume that each of the training, development and test data will contain a mix of both anomalous and non-anomalous log data. These models use the supervised losses like cross entropy loss or squared error loss. But they can suffer due to the under-representativeness of the anomalous class of logs, especially if they occur very rarely in the training and development data. Due to their direct dependency on modeling the anomalous class explicitly these models also lack robustness when the anomaly distribution changes. 

\textbf{Unsupervised Anomaly Detection models} do not require any anomaly label for the log data. But the existing unsupervised models in the literature typically assume that the entire training data is comprised of only normal or non-anomalous logs and generally show a sharp decline in performance when the training data is adulterated with even a small fraction of anomalous logs. Amongst the most popular unsupervised anomaly detection models, mainly two paradigms have been followed: 
\vspace{-.5em}
\begin{itemize}[leftmargin=*]
\item \textbf{Forecasting based models}: These models learn the representations of the log lines through a forecasting based self-supervision i.e. by learning to predict the label of next log line given an input context of log sequence. For all of these models, following Deep-Loglizer paper, the label has been taken as the event id of the next log line. This category of models includes various sequence encoding networks that have been popular in deep-learning - like recurrent neural network or convolutional neural network based models or the more recent, more powerful self-attention based Transformer models. These models are typically trained a cross-entropy loss between the true and predicted distributions, which aims to maximise the likelihood of the true label, conditional to the given input sequence. 
\vspace{-.5em}
\item \textbf{Reconstruction based models}: This includes Auto-encoder based models which try to reconstruct a given sequence of loglines through a learnable hidden layer that learn an n-dimensional representation of each log-line. The other more recent models in this category are Transformer based models which are trained using masked-language modeling principles. During training a certain fraction of the input tokens would be masked and the model would learn to predict these tokens using the remaining input context; and in the process learning the contextual representation of each token in a log-line or a log-sequence. This is the fundamental principle behind BERT Language model with the masked language  modeling providing the learning objective when training on the log data in a self-supervised way. 
\end{itemize}
\vspace{-.5em}
\textbf{Forecasting based Anomaly Detection:} For our benchmarking with forecasting based models, we select three core deep learning models which have been the basis of the some of the most popular recent neural log anomaly detection methods 
\vspace{-.5em}
\begin{itemize}[leftmargin=*]
    \item \textbf{LSTM:} This model corresponds to a long-short term memory (LSTM) network to encode a given log sequence. It also provides various options - i) whether to utilize uni-directional or bi-directional encoding of tokens in a given input sequence ii) whether to have a learnable attention network over the input sequence, which linearly combines the hidden representations with the attention weights. 
    \item \textbf{CNN:} This model corresponds to a convolutional neural network (CNN) to encode a given log sequence. Different convolutional layers with different shape settings are applied on the input followed by a 1-d max-pooling operation. The outputs from each of these are then concatenated and fed into a fully-connected layer. 
    \item \textbf{Transformer:} This model corresponds to a Transformer based encoder network with a multi-headed self-attention mechanism to encode a given log sequence. Since the Transformer outputs a d-dimensional representation for each token in the input log-sequence, a mean-pooling operation is applied over those representations, to get a fixed representation for the entire sequence. 
\end{itemize}

Since the LSTM, CNN and Transformer models need a $d$-dimensional representation of each log, first an embedding layer is applied to the raw log input features. In case of sequential feature representation, each log event id is embedded as a $d$-dimensional vector, while for semantic feature representation, the embedding layer is initialized with the pretrained embeddings (e.g. Word2Vec or FastText etc) and embeds each log token id to a $d$-dimensional vector. 

The output of the LSTM, CNN or Transformer a fixed $d$-dimensional representation of the input sequence which is then downward projected to 1-d space, followed by a softmax layer. For supervised versions of these models, since the explicit label (anomalous or not) exists for each log-line or log-sequence, the output of the softmax layer is aimed to directly predict this label. For forecasting based unsupervised versions, the output of the softmax layer is aimed to predict the id of the next log-line, that is succeeding the given input log sequence. During inference, for forecasting based unsupervised models make a prediction for a given input log sequence, which is then compared against the actual log event following the input sequence. We follow the similar inference strategy as \cite{DBLP:journals/corr/abs-2107-05908} and predict a test instance as anomalous if the ground truth is not one of the $k$ (=10) most probable log events predicted by the model. A smaller $k$ imposes more demanding requirements on the model’s performance.

In literature, LSTM based models have been used by DeepLog \cite{10.1145/3133956.3134015}, LogAnomaly \cite{10.5555/3367471.3367702} and LogRobust \cite{Zhang2019}. While DeepLog uses sequential representations, where each log message is represented by the index of its log event, LogAnomaly uses semantic representations. While both of these use unidirectional LSTM in an unsupervised setting, LogRobust uses supervised version of an bi-directional LSTM with the attention network. CNN has been used by \cite{DBLP:conf/dasc/LuWLW18} but only in a supervised setting. Transformer based model has been applied in LogSy \cite{9338283}, but they additionally use auxiliary log datasets as pseudo-anomalous data. This helps them to learn a better representation of normal log data from the target system of interest while regularizing against overfitting. In order to ensure better reproducibility, in our benchmarking we do not use any additional log datasets and hence in some of the supervised settings, our Transformer based models suffer from overfitting issues and yield somewhat poorer results and are not directly comparable to the results obtained by \cite{DBLP:conf/dasc/LuWLW18}. Following \cite{DBLP:journals/corr/abs-2107-05908} for all of these models, in both the supervised and unsupervised settings, we report the F1-Scores. 

\textbf{Reconstruction based Anomaly Detection:} For our benchmarking with reconstruction based models, we select the LogBERT model from the work LanoBERT \cite{DBLP:journals/corr/abs-2111-09564}. Following that literature, the preprocessing configurations are set before the BERT model can be applied - i) Since LogBERT is a parser-free technique, no log parsing is applied. ii)  For obtaining the vectorized log representation, the preprocessed log sequences are tokenized using the WordPiece (Wu et al. 2016) model used in BERT. iii) The tokenizer is trained from scratch on each log dataset to ensure that the dataset-specific custom vocabulary can be learned. During training the usual masked language modeling principles of BERT is followed. During inference, multiple masked versions of each test instance is generated, by passing a fixed-size masking window over the token sequence, ignoring masking of special characters. Thus a test instance of sequence length $N$ will result in an average of $\frac{N}{n}$ masked instances, each have a masked n-gram of length upto $n$. After running the inference on the masked test instance, the anomaly score is obtained as the average of the top-prediction probabilities (or log-probabilities) over the k-most confident masked tokens. Following LanoBERT, we report AUROC (Area under ROC) metric over this anomaly score. 

All unsupervised models, (forecasting or reconstruction based) are trained only on the normal log data. Following Deep-Loglizer, for the forecasting based models, around 20\% of the data and for LogBERT, following LanoBERT, around 30\% of the data is sequestered for test. These percentages include the entire set of anomalous logs in the dataset. In LogAI, we take out 10\% of the training data as development set for validation and model selection purposes. 

\subsection{Datasets:} Following Deep-Loglizer and LanoBERT, we perform our benchmarking experiments on two of the most popular public log anomaly detection datasets - HDFS and BGL. Additionally for LogBERT we also benchmark on the public dataset, Thunderbird. 
Further, similar to Deep-Loglizer, for BGL dataset we also perform a fixed-window based log partitioning by grouping log-lines over every 6-hourly window.  However for LogBERT model, following LanoBERT, we treat each individual log-line as a train or test instance, without doing any log partitioning. On the other hand, for HDFS dataset, since anomaly labels are available only at the level of each session-id (which is also known as BLOCK in the raw dataset), we use identifier based log partitioning, by constructing log-sequences for each session-id. These resulting log partitions are treated as the training or test instances for all algorithms. 

\subsection{Experimental Settings and Results:} 
\begin{table*}[!htbp]
\centering
\resizebox{\textwidth}{!}{
{\renewcommand{\arraystretch}{1.2}
\begin{tabular}{|l|l|l|l|l|c|c||c|c|} \hline 
 \textbf{Model} & \textbf{Details} & \textbf{Supervision} & \parbox{1.5cm}{\textbf{Log Parsing}} & \parbox{1.5cm}{\textbf{Log Representation}}  & \multicolumn{2}{c||}{\textbf{HDFS}} & \multicolumn{2}{c|}{\textbf{BGL}} \\ \cline{6-9}
  & & & & & \textbf{LogAI} & \parbox{1cm}{\textbf{Deep-Loglizer}} & \textbf{LogAI} & \parbox{1cm}{\textbf{Deep-Loglizer}}  \\ \hline

\multirow{8}{*}{\textbf{LSTM}} & \multirow{4}{*}{\parbox{2.7cm}{Unidirectional, No Attention}} & \multirow{4}{*}{Unsupervised} & \multirow{2}{*}{\cmark} & sequential & 0.981 & 0.944 & 0.938 & 0.961 \\ \cline{5-9}
& & & & semantic &  0.981 & 0.945 & 0.924 & 0.967 \\ \cline{4-9}
& & & \multirow{2}{*}{\xmark} & sequential & 0.979 & - & 0.925 & - \\  \cline{5-9}
& & & & semantic &  0.981 & - & 0.924 & - \\ \cline{2-9}

& \multirow{4}{*}{\parbox{2.7cm}{Bidirectional, With Attention}} & \multirow{4}{*}{Supervised} & \multirow{2}{*}{\cmark} & sequential &  0.984 & 0.96 & 0.983 & 0.983 \\ \cline{5-9}
& & & & semantic &  0.964 & 0.964 & 0.95 & 0.983 \\ \cline{4-9}
& & & \multirow{2}{*}{\xmark} & sequential &  0.989 & - & 0.931 & - \\ \cline{5-9}
& & & & semantic &  0.971 & - & 0.983 & - \\\hline\hline 

\multirow{4}{*}{\textbf{CNN}} &  \multirow{4}{*}{\parbox{3cm}{2-D Convolution with 1-D Max pooling}} & \multirow{2}{*}{Unsupervised} & \cmark & sequential & 0.981 & - & 0.929 & - \\ \cline{4-9}
& & & \xmark & sequential &  0.981 & - & 0.922 & - \\ \cline{3-9}
& & \multirow{2}{*}{Supervised} & \cmark & sequential &  0.943 & 0.97 & 0.983 & 0.972 \\ \cline{4-9}
& & & \xmark & sequential &  0.946 & - & 0.990 & - \\ \hline \hline 

\multirow{8}{*}{\textbf{Transformer}} & \multirow{8}{*}{\parbox{2.7cm}{Multihead single-layer self-attention model, trained from scratch}} & \multirow{4}{*}{Unsupervised} & \multirow{2}{*}{\cmark} & sequential &  0.971 & 0.905 & 0.933 & 0.956 \\ \cline{5-9}
& & & & semantic &  0.978 & 0.925 & 0.921 & 0.957 \\ \cline{4-9} 
& & & \multirow{2}{*}{\xmark} & sequential &  0.98 & - & 0.92 & - \\ \cline{5-9}
& & & & semantic &  0.975 & - & 0.917 & - \\ \cline{3-9}

& & \multirow{4}{*}{Supervised} & \multirow{2}{*}{\cmark} & sequential &   0.934
& - & 0.986 & - \\ \cline{5-9}
& & & & semantic &   0.784 & - & 0.963 & - \\ \cline{4-9} 
& & & \multirow{2}{*}{\xmark} & sequential &  0.945 & - & 0.994 & - \\ \cline{5-9} 
& & & & semantic &  0.915 & - & 0.977 & - \\ \hline 
\end{tabular}}}
\caption{Comparison between different supervised and unsupervised Forecasting-based neural anomaly detection models in LogAI and Deep-Loglizer library \cite{DBLP:journals/corr/abs-2107-05908}, using F1-Score as the performance metric. The dashed (-) cells indicate that there are no reported numbers in the Deep-Loglizer paper corresponding to those configurations.}
\label{tab:nn_results}
\end{table*}
For our benchmarking we conduct experiments on the above choice of anomaly detection algorithms under various settings and compare our experimental results with those published in Deep-Loglizer \cite{DBLP:journals/corr/abs-2107-05908} and LanoBERT \cite{DBLP:journals/corr/abs-2111-09564} papers In Table \ref{tab:nn_results} we list the performance of the different supervised and unsupervised forecasting-based models (LSTM, CNN and Transformer), while \ref{tab:LogBERT_results} shows the results using the unsupervised reconstruction-based LogBERT model.

\textbf{Evaluation Metrics:} In order to compare the performances, for all supervised and unsupervised forecasting-based models we use F1-Score as the metric, following Deep-Loglizer paper. Whereas, for LogBERT, following LanoBERT paper we report the AUROC metric. LanoBERT paper also provides F1-Score, but the F1-Score calculation needs fixing a threshold, which is challenging to do over the training data that only has normal logs. According to the paper, their reported scores are the best F1 value that was calculated using the threshold that yields the best performance for the test dataset. This is not a fair metric, as it involves label-knowledge of the blind test set and hence we only compare using AUROC metric. 

\textbf{Configuration Settings for Evaluation:} For each of LSTM and Transformer models, we benchmark on 8 different configuration settings for each dataset - based on the kind of supervision (supervised or unsupervised), whether log parsing is applied or not, whether the log representation is sequential or semantics based. For CNN models, we found the semantics based log representation results in very slow convergence rate, hence we have benchmarked our results using only the sequential feature representations of the logs. On the other hand, Deep-Loglizer showcases only specific settings for these models - for e.g. forecasting based unsupervised anomaly detection is done using Unidirectional LSTM with no-attention and Transformer network while supervised models are Bidirectional LSTM with attention and CNN network, whereas all of these methods can be applied on both supervised and unsupervised settings. Each of their models use the Log Parsing step and have two variants that use sequential and semantic feature representations for the logs. However Deep-Loglizer paper \cite{DBLP:journals/corr/abs-2107-05908} provides only 8 configurations for each dataset whereas LogAI is benchmarked on a more exhaustive set of 20 configurations per dataset. 

\textbf{Performance Comparison:} In most of these configurations the performance achieved by LogAI is comparable to that of Deep-Loglizer. The 2-3\% difference in performance between the models is not quite statistically significant and can mostly be attributed to the following factors:  Following the implementation open-sourced by authors of Deep-Loglizer in \url{https://github.com/logpai/deep-loglizer}, it is evident that the library does not utilize any development (or validation) set and directly performs model selection based on the test performance. LogAI on the other hand, selects the model checkpoint on the validation performance and reports the results on the blind test set. Secondly, because of the same reason the resulting the training and test splits used by LogAI and Deep-Loglizer are not identical. Especially for BGL data, perhaps the performance difference is somewhat more observeable, since both libraries apply fixed time-partitions of 6 hours and reports the evaluation at the level of the partitions, instead of the logline level evaluation. This also adds to the possibility of more significant differences in the training/test data setup between the two models. 

For Transformer based models, especially in the supervised setting, we observe a reduced performance. Similar effect had been studied in the original work \cite{9338283} that used Transformer model as Log Anomaly Detector in the supervised setting. Their model suffered from overfitting on the target system's log data due to the presence of only rare and sparse anomaly patterns in the train data. To overcome the overfitting issue, they additionally involve other external system's logs as auxiliary data - treating them as pesudo ``anomalous'' logs. But in order to keep our benchmarking reproducible, we do not use any additional auxiliary data abd subsequently report a poorer performance. The Deep-Loglizer paper also benchmarks with only the unsupervised setting of the Transformer model, which is much less prone to overfitting. 

For LogBERT model, we benchmark the test results taking various inferencing strategies. Given a test instance, which has been converted to multiple masked versions (each having a continuous n-gram masked), either we average the inference score either over all the masked tokens or over the top-6 most confident ones, based on the the model prediction likelihood. For the latter we consider different inference scores - mean predictive loss or maximum predictive probability or log probability or the entropy of the prediction distribution. All of these metrics are quite correlated and our objective is to simply show that our LogBERT implementation yields reasonably stable results across these different inferencing strategies. While LanoBERT also uses Predictive Loss and Probability based scores, they provide AUROC evaluation metric metric only for the latter and they also evaluate only HDFS and BGL dataset. In the predictive probability based inference strategy, results obtained by LogAI and LanoBERT are quite comparable, with small differences owing to the variability of the train, test splits used in the two implementations (The authors of LanoBERT have used their own train test split due to the general lack of standardized data splits for these datasets).

\begin{table*}
\centering
\resizebox{0.8\textwidth}{!}{
{\renewcommand{\arraystretch}{1.2}
\begin{tabular}{|l|c|c||c|c||c|c|} \hline 
\textbf{Inference Strategy} & \multicolumn{6}{|c|}{\textbf{Datasets}}  \\ \cline{2-7} 
& \multicolumn{2}{|c||}{\textbf{HDFS}} & \multicolumn{2}{c||}{\textbf{BGL}} & \multicolumn{2}{c|}{\textbf{Thunderbird}} \\ \cline{2-7} 
& \textbf{LogAI} & \textbf{LanoBERT} & \textbf{LogAI} & \textbf{LanoBERT }& \textbf{LogAI} & \textbf{LanoBERT} \\ \hline \hline 
\multicolumn{7}{|c|}{\textbf{Averaged over all masked tokens }} \\ \hline 
\textbf{Mean Predictive Loss} & 0.983 & - & 0.998 & - & 0.953 & - \\ \hline\hline  
\multicolumn{7}{|c|}{\textbf{Averaged over top-6 most-confident masked tokens}} \\ \hline 
\textbf{Mean Predictive Loss} & 0.98 & - & 0.964 & - & 0.937 & - \\ \hline 
\textbf{Max Predictive Prob.} & 0.976 & 0.99 & 0.972 & 0.972 & 0.953 & - \\ \hline 
\textbf{Max Predictive LogProb.} & 0.976 & - & 0.969 & - & 0.917 & - \\ \hline 
\textbf{Mean Predictive Entropy} & 0.976 & - & 0.973 & - & 0.967 & - \\ \hline 
\end{tabular}}}
\caption{Comparison of LogBERT model performance achieved by our LogAI library and by LanoBERT \cite{DBLP:journals/corr/abs-2111-09564}, using the AUROC metric. Both versions of the model are in unsupervised setting (trained on normal logs only) and do not need any log parsing. The dashed (-) cells indicate that there are no reported numbers in the LanoBERT paper corresponding to those configurations.}
\label{tab:LogBERT_results}
\end{table*}

Overall our experiments on the suite deep learning based log anomaly detection models suggests that their implementations in the LogAI library is able to reproduce the established performance benchmarks on standard open-source datasets with reasonable accuracy. Additionally, owing to a more generic data processing pipeline we are seamlessly able to extend to a more exhaustive set of experimental settings, than what has been explored or implemented before in existing literature and libraries. 

\section{Conclusion}
\label{sec:conclusion}
In this technical report we introduced LogAI, an open source library for AI-based log analytics and intelligence. LogAI library uses the same unified log data model as OpenTelemetry to ensure the analytical processes to be agnostic to any log platforms that supports OpenTelemetry. LogAI also abstracts common processes in different downstream tasks and provides reusable components to execute these processes. LogAI also provides a large varieties of AI capabilities, from time-series anlaysis, traditional statistical learning to deep learning and pre-trained transformer models. We showed how LogAI can be used to conduct a variety of common log analysis tasks such as log summarization, clustering and anomaly detection and also provide extensive benchmarking results on Log Anomaly Detection. LogAI version v0.1.0 is released as open-source code under BSD-3-Clause license. Our team will provide continuous support and further improvements in the future versions.

\section*{Acknowledgments}
We would like to thank a number of leaders and colleagues from Salesforce.com Inc. who have provided strong support, advice, and contributions to this open-source project.

\bibliographystyle{unsrt}  
\bibliography{templateArxiv}

\end{document}